\documentclass[letterpaper, 10 pt, conference]{ieeeconf}  

\IEEEoverridecommandlockouts                              

\usepackage{caption}
\usepackage{xcolor}
\usepackage{multirow}
\usepackage{makecell}    
\usepackage{dsfont}  
\usepackage{graphicx}
\usepackage{float} 
\usepackage{subfigure}
\usepackage{amsmath}
\usepackage{amssymb}
\usepackage{booktabs}
\usepackage{graphicx}
\usepackage{wrapfig}
\usepackage{lipsum} 
\usepackage{hyperref} 
\usepackage{url}
\usepackage{bm}

\hypersetup{
    colorlinks=true,
    linkcolor=black,
    filecolor=black,      
    urlcolor=blue,
    citecolor=black,
}

\title{\LARGE \bf
Contrastive Representation Learning for Robust Sim-to-Real Transfer of Adaptive Humanoid Locomotion
}

\author{
    Yidan Lu$^{1,2*}$, Rurui Yang$^{2*}$, Qiran Kou$^{2*}$, Mengting Chen$^{2\dagger}$,\\
    Tao Fan$^{2}$, Peter Cui$^{2}$, Yinzhao Dong$^{1}$, and Peng Lu$^{1\S}$%
    \thanks{$^{*}$ Equal Contributions. $^{\dagger}$ Project leader. $^{\S}$ Corresponding author.}%
    \thanks{$^{1}$ Adaptive Robotic Controls Lab (ArcLab), Department of Mechanical Engineering, The University of Hong Kong, Hong Kong SAR, China. Yidan Lu: \url{ydlu@connect.hku.hk}; Yinzhao Dong: \url{dongyz@connect.hku.hk}; Peng Lu: \url{lupeng@hku.hk}}%
    \thanks{$^{2}$ PNDbotics, China. \url{rurui.yang@pndbotics.com}; \url{qiran.kou@pndbotics.com}; \url{mengting.chen@pndbotics.com}; \url{tao.fan@pndbotics.com}; \url{peter.cui@pndbotics.com}}%
    \thanks{This work was supported by the General Research Fund under Grant 17204222. This work was conducted during Yidan Lu's internship at PNDbotics. The authors would like to thank PNDbotics for providing the hardware platform and technical support.}%
}

\begin{document}
\maketitle
\thispagestyle{empty}
\pagestyle{empty}

\begin{abstract}
Reinforcement learning has produced remarkable advances in humanoid locomotion, yet a fundamental dilemma persists for real-world deployment: policies must choose between the robustness of reactive proprioceptive control or the proactivity of complex, fragile perception-driven systems. This paper resolves this dilemma by introducing a paradigm that imbues a purely proprioceptive policy with proactive capabilities, achieving the foresight of perception without its deployment-time costs. Our core contribution is a contrastive learning framework that compels the actor's latent state to encode privileged environmental information from simulation. Crucially, this ``distilled awareness" empowers an adaptive gait clock, allowing the policy to proactively adjust its rhythm based on an inferred understanding of the terrain. This synergy resolves the classic trade-off between rigid, clocked gaits and unstable clock-free policies. We validate our approach with zero-shot sim-to-real transfer to a full-sized humanoid, demonstrating highly robust locomotion over challenging terrains, including 30 cm high steps and 26.5° slopes, proving the effectiveness of our method. Website:~\url{https://lu-yidan.github.io/cra-loco}.
\end{abstract}

\section{INTRODUCTION}

Achieving stable and adaptive locomotion in unstructured environments is a grand challenge for humanoid robotics. While Deep Reinforcement Learning (DRL) has become a cornerstone for synthesizing such behaviors, a fundamental information gap complicates real-world deployment. Policies are typically trained in simulation where they have access to privileged information—such as precise terrain geometry, friction coefficients, and robot dynamics. However, upon deployment, they must operate using only onboard proprioceptive sensors (e.g., joint encoders, IMU), losing all direct knowledge of the external world. This gap forces a difficult choice between two dominant strategies for achieving robustness.

The first strategy is to train a robustly reactive policy. Through extensive domain randomization, the policy learns to react to disturbances inferred from proprioceptive feedback~\cite{siekmann2021sim,siekmann2021blind,xie2020learning,xie2018feedback,rodriguez2021deepwalk,li2021reinforcement,li2023robust,li2024reinforcement,duan2022learning, radosavovic2024real}. While highly robust, these policies are fundamentally ``blind''; they can withstand an unexpected push but cannot proactively adjust for an upcoming obstacle, limiting their performance on highly challenging terrain. The second strategy is to equip the robot with exteroceptive sensors like cameras or LiDAR, allowing the policy to directly perceive the environment and plan ahead~\cite{duan2024learning, zhuang2024humanoid, allshire2025visual, wang2025moremixtureresidualexperts}. This enables proactive control but introduces significant system complexity, computational overhead, and potential points of failure in the perception pipeline.

\begin{figure}[t]
    \centering
    \includegraphics[width=0.5\textwidth]{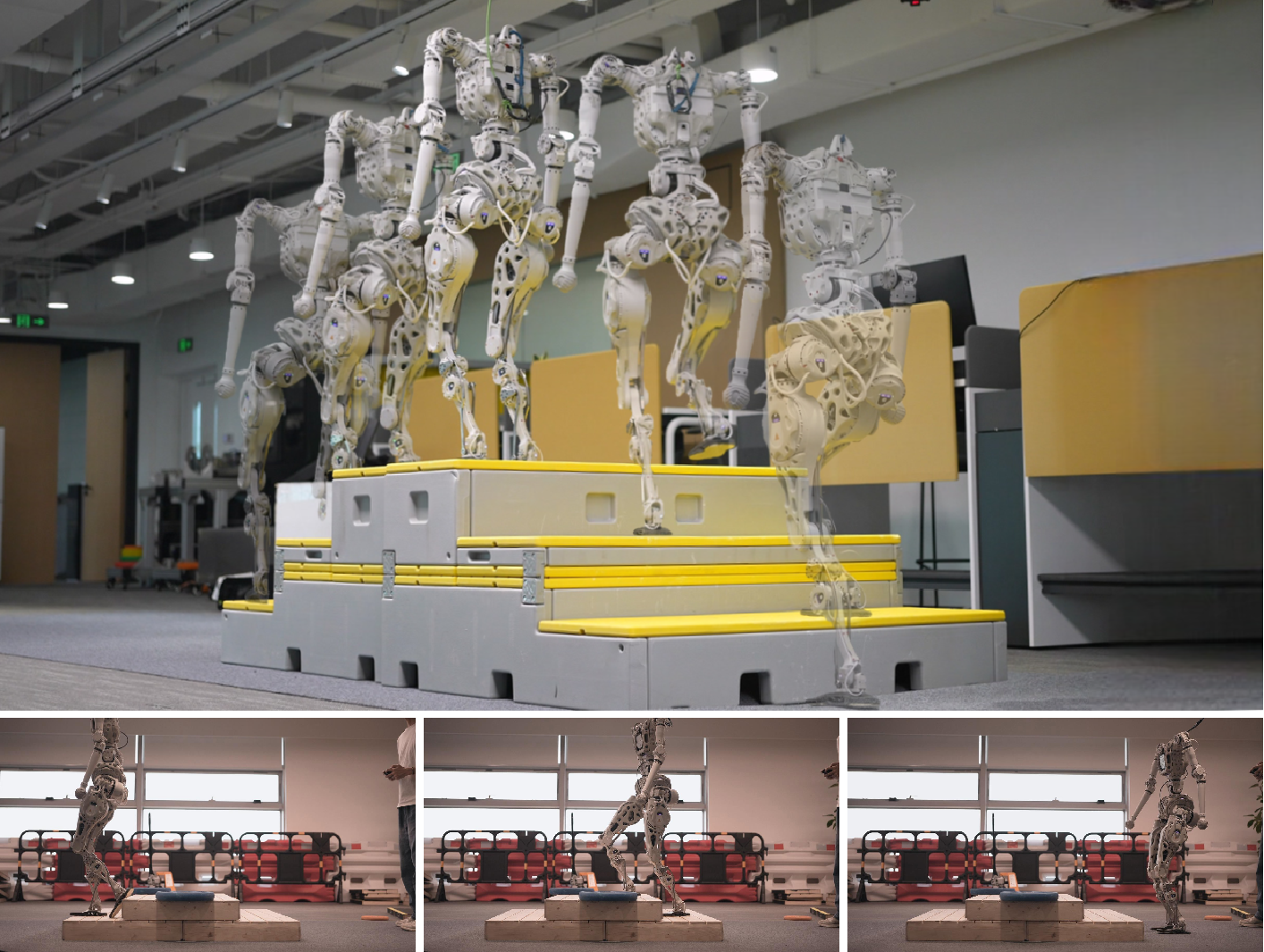}
    \caption{Our policy, trained via contrastive knowledge distillation, enables the Adam Lite humanoid~\cite{adamlite} to achieve robust zero-shot sim-to-real locomotion over challenging terrains. Top: Ascending a 30 cm high staircase. Bottom: A motion sequence of climbing a 10 cm staircase.}
    \label{fig:frame}
\end{figure}

In this paper, we introduce a paradigm that transcends the reactive nature of traditional proprioceptive control, achieving proactive adaptation without the deployment-time cost of exteroception. Our approach is centered on \textbf{contrastive knowledge distillation} within an asymmetric actor-critic framework. The key idea is to force the actor's proprioceptive state representation to become predictive of the privileged environmental context that is available only to the critic during simulation. By training the actor to distinguish ``matching'' pairs of proprioceptive history and environmental context from ``non-matching'' pairs, our contrastive loss encourages the policy's latent state to encode terrain-relevant information. The policy thereby learns to infer environmental properties by associating subtle patterns in its own dynamic response with their underlying causes.

A policy endowed with such an inferred environmental awareness is uniquely positioned to leverage control structures that offer dynamic flexibility. We demonstrate this by integrating our learning framework with an adaptive gait mechanism, where the policy can dynamically modulate its gait frequency and phase. While the ability to adapt gait is not new~\cite{duan2024learning, ijspeert2008central, li2024ai, zhang2021adaptive}, our method provides the crucial missing piece: the \textit{wisdom} for the policy to know \textit{when} and \textit{how} to adapt based on its internalized, data-driven understanding of the environment. This synergy creates a controller that is efficient, proactive, and lightweight.

Our work makes the following contributions:
\begin{itemize}
    \item A novel training framework that uses contrastive learning to distill an awareness of privileged environmental properties into a purely proprioceptive policy, bridging the sim-to-real information gap.
    \item A novel method that resolves the classic trade-off between rigid clocked gaits and inefficient clock-free policies, by using the distilled awareness to intelligently inform an adaptive gait clock.
    \item Comprehensive real-world validation on a full-sized humanoid, demonstrating highly robust, zero-shot sim-to-real locomotion over challenging terrains like 30 cm steps and 26.5° slopes, confirming the practical effectiveness of our approach.
\end{itemize}


\section{RELATED WORK}

Classical approaches to bipedal locomotion are predominantly model-based, relying on principles like the Zero-Moment-Point (ZMP)~\cite{kajita2003biped} and simplified dynamics such as the Linear Inverted Pendulum Model (LIPM)~\cite{kajita20013d}. While extensions using Model Predictive Control (MPC) have enhanced robustness~\cite{wieber2006trajectory, englsberger2013three}, these methods are fundamentally constrained by model accuracy and require meticulous parameter tuning. Their fragility in unstructured environments~\cite{xie2021compliant} has catalyzed the widespread adoption of learning-based techniques.

Deep Reinforcement Learning (DRL) has shown remarkable success in synthesizing complex locomotion skills, which map joint states and IMU readings to actions. These methods achieved impressive robustness, often by leveraging strong priors like reference motions~\cite{xie2018feedback, li2021reinforcement, peng2017deeploco} or dedicated phase cycles~\cite{siekmann2021sim, gu2024humanoid,xue2025unified}. While some approaches could learn to traverse varied terrain without explicit guidance~\cite{duan2022learning,radosavovic2024real}, their adaptability largely stemmed from extensive domain randomization during training, making them highly reactive to unforeseen contacts rather than proactive to upcoming terrain. This reactive nature fundamentally limits their performance on highly discontinuous or challenging ground where foresight is critical.

To imbue policies with foresight, the research frontier has shifted towards integrating exteroceptive perception. A dominant trend is the use of onboard cameras or radars, which could help provide rich environmental context, enabling remarkable feats like traversing highly challenging obstacles~\cite{duan2024learning, zhuang2024humanoid, long2025learning, he2025attention} and navigating sparse footholds~\cite{wang2025beamdojo}. This paradigm has led to state-of-the-art real-world humanoid deployments and unified whole-body controllers~\cite{allshire2025visual}. However, this performance is predicated on the availability and reliability of deployment-time perception hardware and complex processing pipelines, introducing potential points of failure.


An emerging paradigm seeks to resolve this tension by distilling privileged simulation knowledge into a proprioceptive policy, effectively teaching it to ``see" without eyes. Early approaches employed a teacher-student framework, where a privileged ``teacher" policy trained with full information guides a proprioceptive ``student"~\cite{kumar2022adapting}. However, this introduces multi-stage complexity and inherently caps the student's performance. To overcome this, recent work has shifted to end-to-end representation learning, which can be broadly categorized. One path involves auxiliary prediction, training a world model to explicitly reconstruct privileged information (e.g., terrain maps) from proprioception~\cite{Gu-RSS-24, sun2025learning}. While direct, this adds significant overhead from training a separate, complex predictive model. Another path employs contrastive learning, but often with a \textit{temporal} objective—aligning historical and future observations to learn predictive dynamics~\cite{long2023hybrid, long2025learning}.

Our work introduces a more direct and efficient contrastive approach. Instead of learning temporal dynamics or reconstructing the full state, we employ a \textbf{spatial contrastive objective} that directly aligns the policy's latent representation with the immediate environmental context. This integrated, end-to-end method bypasses the inefficiency of auxiliary world models and the complexity of teacher-student frameworks. By distilling spatial context, our approach informs the entire policy's representation, enabling proactive adaptation. The synergy between this informed policy and a flexible control structure like an adaptive clock resolves the classic trade-off between rigid clocked gaits and inefficient clock-free policies, resulting in a lightweight, robust, and proactive controller.
\section{Methods}
\label{sec:method}

\begin{figure}[t]
    \centering
    \includegraphics[width=0.5\textwidth]{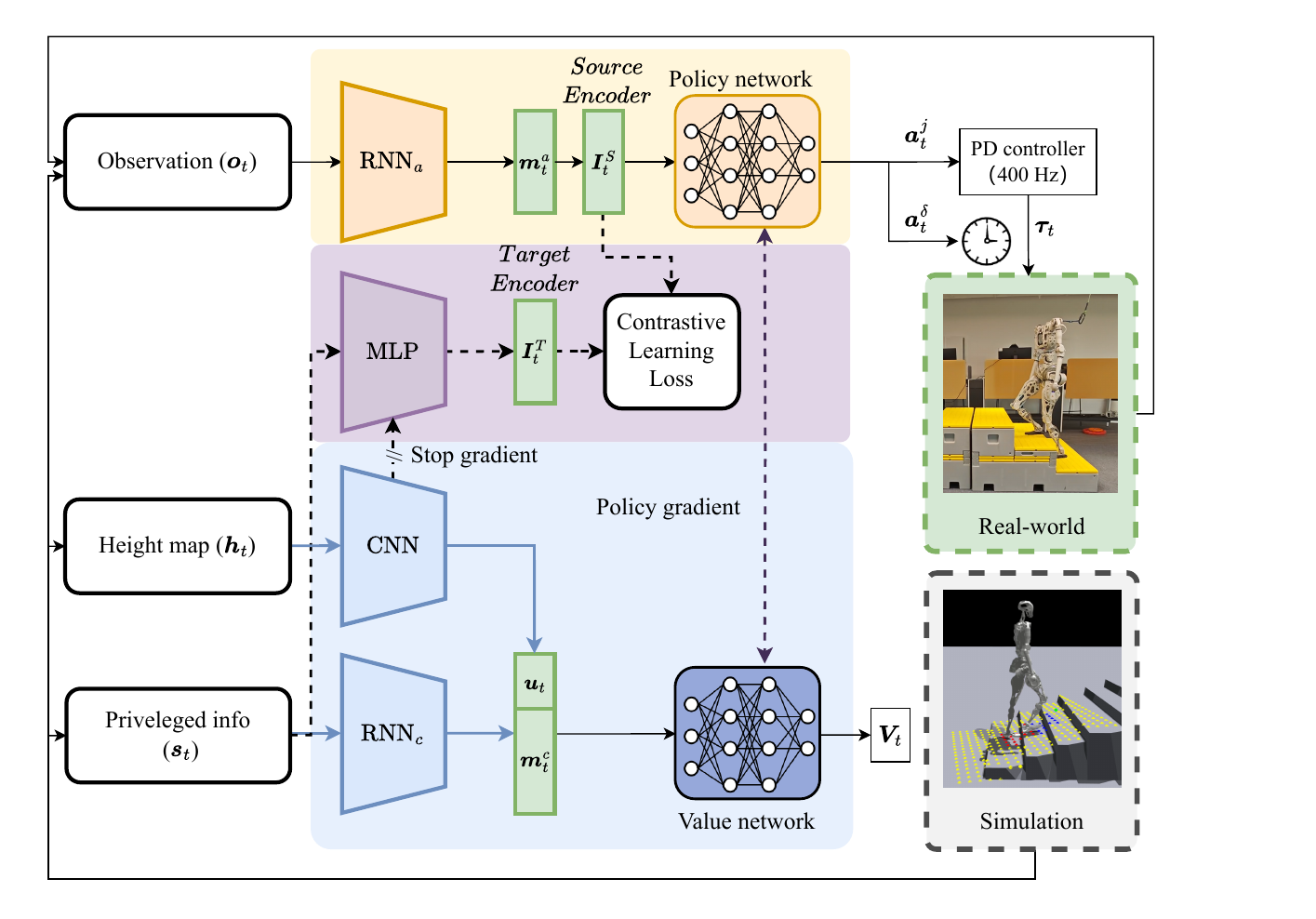}
    \caption{Overview of our proposed training framework. An asymmetric Actor-Critic structure is employed: the Actor relies on local sensor feedback for decision making, while the Critic incorporates additional privileged information (e.g., terrain height maps) to provide more accurate value estimation. Both the Actor and Critic networks use 256-dimensional intermediate representations from RNN modules.}
    \label{fig:framework}
\end{figure}

As shown in Fig.~\ref{fig:framework}, we propose an \textbf{Adaptive Gait} framework with two main networks: an \textbf{Actor} and a \textbf{Critic}, trained via an asymmetric Actor-Critic approach. The Actor relies on partial observations (e.g., onboard sensor data) for real-world deployment, while the Critic exploits additional privileged information (such as a height map) during simulation. Both networks use recurrent neural networks (RNNs) to capture temporal dependencies, each providing a 256-dimensional hidden state for subsequent layers. In the following sections, we describe how the Actor’s 84-dimensional sensor input is transformed into a 26-dimensional action, and how the Critic processes augmented observations, together with CNN-extracted height-map features, to accurately estimate returns. 

\subsection{Asymmetric Actor-Critic Network}
Humanoid locomotion in unstructured environments can be naturally formulated as a Partially Observable Markov Decision Process (POMDP) \cite{spaan2012partially}, where the agent only has access to partial or noisy observations of the full environment state. To address these limitations, we adopt an asymmetric actor-critic design: the actor receives a compact sensor-based representation for decision making, while the critic is given more comprehensive data (often accessible only during training) for accurate value estimation. Our results show that combining these distinct information streams yields robust terrain-aware locomotion in high-dimensional continuous control.

\subsubsection{Actor Network}
We define the actor network $\pi_\theta(\boldsymbol{a}_t \mid \boldsymbol{m}_t^a)$, parameterized by $\theta$, to output an action $\boldsymbol{a}_t \in \mathbb{R}^{26}$ given an intermediate observation $\boldsymbol{m}_t^a \in \mathbb{R}^{256}$ from \(\mathrm{RNN}_a\). Each raw sensor observation $\boldsymbol{o}_t \in \mathbb{R}^{84}$ comprises the control command $\boldsymbol{c}_t \in \mathbb{R}^3$, projected gravity $\boldsymbol{g}_t \in \mathbb{R}^3$, joint angles $\boldsymbol{q}_t \in \mathbb{R}^{23}$ and velocities $\dot{\boldsymbol{q}}_t \in \mathbb{R}^{23}$, previous action $\boldsymbol{a}_{t-1} \in \mathbb{R}^{26}$, body angular velocity $\boldsymbol{\omega}_t \in \mathbb{R}^3$, average angular speed $\boldsymbol{\overline{\omega}}_t \in \mathbb{R}^1$, and gait phase $\boldsymbol{\psi}_t \in \mathbb{R}^2$. An RNN (\(\mathrm{RNN}_a\)) processes $\boldsymbol{o}_t$ along with its hidden state $\boldsymbol{h}_{t-1}^{\mathrm{actor}} \in \mathbb{R}^{256}$:
\begin{equation}
    \boldsymbol{h}_t^{\mathrm{actor}},\, \boldsymbol{m}_t^a
    =
    \mathrm{RNN}_a(\boldsymbol{o}_t,\;\boldsymbol{h}_{t-1}^{\mathrm{actor}}),
    \label{eq:rnn_actor}
\end{equation}
where $\boldsymbol{h}_t^{\mathrm{actor}} \in \mathbb{R}^{256}$ is the updated hidden representation, and $\boldsymbol{m}_t^a \in \mathbb{R}^{256}$ is fed to a small multilayer perceptron (MLP) to produce the final action $\boldsymbol{a}_t \in \mathbb{R}^{26}$. We split $\boldsymbol{a}_t$ into $\boldsymbol{a}^j_t \in \mathbb{R}^{23}$ and $\boldsymbol{a}^\delta_t \in \mathbb{R}^3$. The term $\boldsymbol{a}^j_t$ represents incremental changes from a reference pose $\boldsymbol{\mathring{q}}$, yielding $\boldsymbol{q}^*_t = \boldsymbol{\mathring{q}} + \boldsymbol{a}^j_t$, while $\boldsymbol{a}^\delta_t$ adjusts the gait period and phase offset. A PD controller then computes the joint torques:
\begin{equation}
  \boldsymbol{\tau} 
  = 
  \boldsymbol{k}_p 
  \bigl(
    \boldsymbol{q}^*_t - \boldsymbol{q}_t
  \bigr)
  + 
  \boldsymbol{k}_d 
  \bigl(
    -\dot{\boldsymbol{q}}_t
  \bigr),
  \label{eq:pd_control}
\end{equation}
where $\boldsymbol{q}_t$ and $\dot{\boldsymbol{q}}_t$ are the current joint angles and velocities, ensuring each joint smoothly tracks the desired configuration.
\subsubsection{Critic Network with Height Map}

The \textbf{Critic} network takes an augmented observation \(\boldsymbol{s}_t\) (including velocity \(\boldsymbol{v}_t\), center-of-mass position, and two \(5\times5\) foot-level height maps) alongside additional privileged signals only available in simulation. A global terrain map \(\boldsymbol{h}_t \in \mathbb{R}^{17\times 11}\) is processed by a CNN to yield a 32-dimensional feature vector \(\boldsymbol{u}_t\). Meanwhile, an RNN operates on \(\boldsymbol{s}_t\):
\begin{equation}
\label{eq:rnn_critic}
  \boldsymbol{h}_t^{\mathrm{critic}},\, \boldsymbol{m}_t^c 
  =
  \mathrm{RNN}_c\bigl(\boldsymbol{s}_t,\;\boldsymbol{h}_{t-1}^{\mathrm{critic}}\bigr),
\end{equation}
producing a 256-dimensional feature \(\boldsymbol{m}_t^c\). We then form
\begin{equation}
  \boldsymbol{z}_t 
  = 
  \bigl[\,
    \boldsymbol{m}_t^c,\, 
    \boldsymbol{u}_t
  \bigr]^\top,
  \label{eq:zt}
\end{equation}
a 288\(\times 1\) column vector that is passed into a fully connected layer to compute \(V(\boldsymbol{s}_t,\boldsymbol{a}_t)\). Combining the RNN’s 256-dimensional representation with the CNN’s 32-dimensional global-terrain encoding enables the Critic to capture both local body dynamics and broader environmental context for more accurate value estimation. 
\subsubsection{Contrastive Representation Learning}
\label{sec:contrastive}

Inspired by the Hybrid Internal Model (HIM)~\cite{long2023hybrid}, we add a contrastive module (purple in Fig.~\ref{fig:framework}) that brings the actor’s RNN hidden state and the critic’s privileged embedding from the \emph{same} time step closer, while pushing embeddings from other steps apart with an InfoNCE loss.  Whereas HIM contrasts a history window $\boldsymbol{o}_{t-H:t}$ against the $\boldsymbol{o}_{t+1}$ to model future disturbances, our same-step actor–critic contrast directly distils privileged terrain cues into the policy.

Let the \textbf{Source Encoder} map the actor’s hidden state \(\boldsymbol{m}_t^a \in \mathbb{R}^{256}\) into \(\boldsymbol{I}_t^S \in \mathbb{R}^{256}\):
\begin{equation}
    \boldsymbol{I}_t^S 
    \;=\;
    \mathrm{MLP}_S\bigl(\boldsymbol{m}_t^a\bigr),
\end{equation}
and let the \textbf{Target Encoder} produce \(\boldsymbol{I}_t^T \in \mathbb{R}^{256}\) by
\begin{equation}
    \boldsymbol{I}_t^T
    \;=\;
    \mathrm{MLP}_T\Bigl(
      \mathrm{stop\_grad}(\boldsymbol{u}_t),\,\boldsymbol{s}_t
    \Bigr),
\end{equation}
where \(\boldsymbol{u}_t\) is the CNN feature (with gradients stopped) and \(\boldsymbol{s}_t\) is the privileged observation. For each step \(t\), \((\boldsymbol{I}_t^S,\;\boldsymbol{I}_t^T)\) forms a positive pair, contrasted against embeddings from other steps. The loss is given by
\begin{equation}
\label{eq:contrast_loss}
   L_{\mathrm{contrast}}
   \;=\;
   -\,\frac{1}{B}\,\sum_{t=1}^{B}
   \log 
   \frac{\exp\bigl(\boldsymbol{I}_t^S \cdot \boldsymbol{I}_t^T / \tau\bigr)}
        {\sum_{t'} \exp\bigl(\boldsymbol{I}_t^S \cdot \boldsymbol{I}_{t'}^T / \tau\bigr)},
\end{equation}
where \(\tau>0\) is a temperature. Minimizing Eq.~(\ref{eq:contrast_loss}) drives embeddings from the same step close in latent space while separating those from different steps, unifying onboard and privileged features into a shared space for more adaptive and terrain-aware locomotion control.
\subsection{Training with PPO}
\label{sec:PPO}

We employ Proximal Policy Optimization (PPO)~\cite{schulman2017proximal}, which constrains each new policy \(\pi_\theta\) near the old policy \(\pi_{\theta_{\text{old}}}\) via a clipped objective:
\begin{equation}
\label{eq:clip_obj}
  L^\mathrm{CLIP}(\theta)
  =
  \mathbb{E}_t\Bigl[
    \min\Bigl(
      r_t(\theta)\,A_t,\; 
      \mathrm{clip}\bigl(r_t(\theta),\,1-\varepsilon,\,1+\varepsilon\bigr)\,A_t
    \Bigr)
  \Bigr].
\end{equation}
where \(r_t(\theta)\) is the ratio of new-to-old action probabilities, \(A_t\) the advantage, and \(\varepsilon\) a small hyperparameter (e.g.\ 0.2). A value loss and entropy regularization further stabilize training and encourage exploration. By limiting the Kullback-Leibler (KL) divergence between consecutive policies, PPO ensures updates remain conservative.

During each rollout, the Actor relies on an 84-dimensional sensor vector and its RNN to produce actions, collecting rewards \(r_t\). The Critic leverages \(\boldsymbol{s}_t\) plus the global map \(\boldsymbol{h}_t\) for accurate returns \(V(\boldsymbol{s}_t,\boldsymbol{a}_t)\). This \textbf{asymmetric} scheme—Critic with privileged data, Actor with onboard sensing—boosts learning efficiency in simulation while preserving deployability on physical platforms.
\subsection{Reward Design}

\begin{figure}[t]
    \centering
    \includegraphics[width=0.48\textwidth]{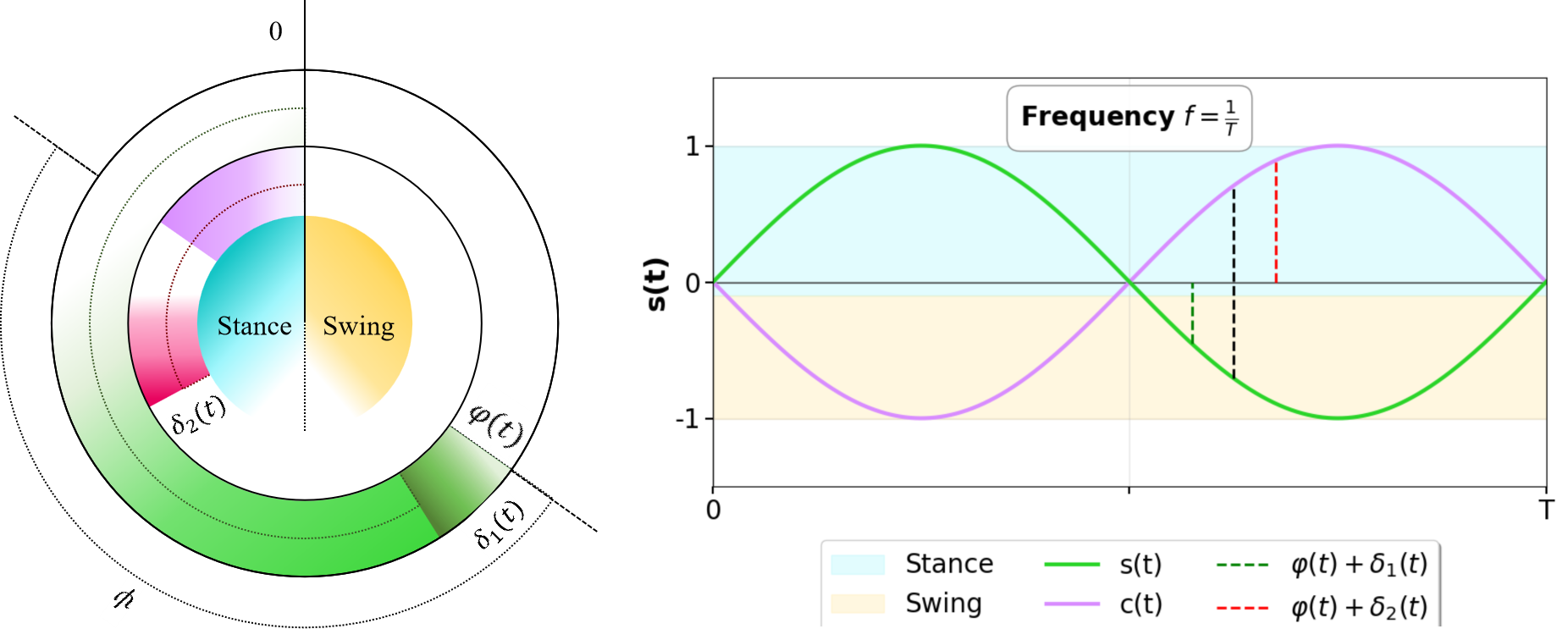}
    \caption{%
      \textbf{Left:} A circular representation of global phase $\phi(t)$,
      showing how $\delta_0(t)$ modifies its rotational speed 
      and $\delta_{1,2}(t)$ shift the leg phases.
      \textbf{Right:} Time-domain view of $s(t)$ (green) and $c(t)$ (purple),
      with stance/swing color bands indicating how phase offsets produce
      distinct footfall timings.%
    }
    \label{fig:gait}
\end{figure}

Our reward function directs the robot to adhere to velocity commands, maintain balance with proper posture, sustain a stable gait, and achieve smooth contact. The reward function is summarized in Table \ref{table:rewards}. It is important to note that the contact pattern reward encourages feet to align with their contact masks, denoting swing and stance phases, as illustrated in Fig.~\ref{fig:gait}. We formulate this as an exponential penalty for any mismatch between the target contact mask, $\boldsymbol{m}_\mathrm{stance}(t)$, and the robot's actual binary contact state, $\boldsymbol{c}_\mathrm{actual}(t)$. This incentivizes the policy to synchronize its physical footfalls with the rhythm dictated by the adaptive clock. The total reward at any time step $t$ is computed as the weighted sum of individual reward components, expressed as $r_t = \sum_i r_i \cdot w_i$, where $w_i$ represents the weighting factor for each reward component $r_i$.

\begin{table}[t]
\caption{Rewards}
\label{table:rewards}
\renewcommand{\arraystretch}{1.2}
\fontsize{6.0pt}{7.0pt}\selectfont
\centering
\begin{tabular}{lll}
\hline
Reward & Equation ($r_i$) & Weight ($w_i$) \\
\hline
Lin. velocity tracking & $\exp \left\{ -\frac{\|\mathbf{v}^{cmd}_{xy} - \mathbf{v}_{xy}\|^2}{\sigma} \right\}$ & $3.0$ \\
Ang. velocity tracking & $\exp \left\{ -\frac{(\omega^{cmd}_{yaw} - \omega_{yaw})^2}{\sigma} \right\}$ & $3.0$ \\
Linear velocity ($z$) & $v_z^2$ & $-0.5$ \\
Angular velocity ($xy$) & $\|\boldsymbol{\omega}_{xy}\|^2$ & $-0.025$ \\
Orientation & $\|\mathbf{g}_{xy}\|^2$ & $-1.25$ \\
Torso Orientation &  $\exp \left\{ -\frac{\|\mathbf{g}_{\text{torso},xy}\|^2}{\sigma} \right\}$  & $2.0$ \\
Joint accelerations & $\|\ddot{\mathbf{q}}\|^2$ & $-2.5 \times 10^{-7}$ \\
Joint power & $\frac{|\boldsymbol{\tau}|^\top|\dot{\boldsymbol{\theta}}|}{\|\mathbf{v}\|^2+0.2\|\boldsymbol{\omega}\|^2}$ & $-2.5 \times 10^{-5}$ \\
Body height w.r.t. feet & $(h^{target} - h)^2$ & $0.1$ \\
Feet clearance & $\sum\limits_{feet} (p_z^{target} - p_z^f)^2 \cdot v^f_{xy}$ & $-0.25$ \\
Action rate & $\|\mathbf{a}_t - \mathbf{a}_{t-1}\|^2$ & $-0.01$ \\
Smoothness & $\|\mathbf{a}_t - 2\mathbf{a}_{t-1} + \mathbf{a}_{t-2}\|^2$ & $-0.01$ \\
Feet stumble & $1\{\exists i, \|\mathbf{F}^{xy}_i\| > 3 F^z_i\}$ & $-0.5$ \\ 
Torques & $\sum\limits_{all~joints} \left(\frac{\tau_i}{kP_i}\right)^2$ & $-2.5 \times 10^{-6}$ \\
Joint velocity & $\sum\limits_{all~joints} \dot{\theta}_i^2$ & $-1 \times 10^{-4}$ \\
Joint tracking error & $\sum\limits_{all~joints} (\theta_i - \theta_i^{target})^2$ & $-0.25$ \\
Arm joint deviation & $\sum\limits_{arm~joints} (\theta_i - \theta_i^{default})^2$ & $-0.1$ \\
Hip joint deviation & $\sum\limits_{hip~joints} (\theta_i - \theta_i^{default})^2$ & $-0.5$ \\
Waist joint deviation & $\sum\limits_{waist~joints} (\theta_i - \theta_i^{default})^2$ & $-1.5$ \\
Joint pos limits & $\sum\limits_{all~joints} out_i$ & $-2.0$ \\
Joint vel limits & $\sum\limits_{all~joints} ReLU(\dot{\theta}_i - \dot{\theta}_i^{max})$ & $-0.1$ \\
Torque limits & $\sum\limits_{all~joints} ReLU(\tau_i - \tau_i^{max})$ & $-0.1$ \\
No fly & $1\{$only one foot on ground$\}$ & $0.25$ \\
Feet lateral distance & $|d^R_{left~foot} - d^R_{right~foot}| - d_{min}$ & $1.0$ \\
Feet slip & $\sum\limits_{feet} |v^{foot}_i| \cdot 1_{\text{new contact}}$ & $-0.05$ \\ 
Feet ground parallel & $\sum\limits_{feet} \text{Var}(H_i)$ & $-2.0$ \\
Feet contact force & $\sum\limits_{feet} ReLU(F^z_i - F_{th})$ & $-1.0 \times 10^{-2}$ \\
Feet parallel & $\text{Var}(D)$ & $-2.5$ \\
Contact momentum & $\sum\limits_{feet} |v^z_i \cdot F^z_i|$ & $-2.5 \times 10^{-2}$ \\ 
Contact pattern & $\exp\left(-\|\boldsymbol{m}_{\text{stance}}(t) - \boldsymbol{c}_{\text{actual}}(t)\|^2\right)$ & $1.0$ \\
\hline
\end{tabular}
\end{table}

\subsection{Adaptive Gait Mechanism Using Additional Action Dimensions}
\label{sec:adaptive_gait}

To enable real-time adaptation of the gait cycle, we augment the action vector $\boldsymbol{a}_t$ with three extra components, $\delta_0(t), \delta_1(t), \delta_2(t)$. While the original components primarily command joint-level poses, these additional dimensions control \emph{gait frequency} (via $\delta_0$) and \emph{foot-phase offsets} (via $\delta_1, \delta_2$). By varying these parameters each timestep, the policy can produce versatile bipedal walking patterns ranging from slow rhythmic motions to faster, more dynamic steps.

\subsubsection{Phase Adaptation}
Let $\Delta(t) = \bigl[\delta_0(t),\,\delta_1(t),\,\delta_2(t)\bigr]^\top$ represent the last three entries of $\boldsymbol{a}_t$, optionally scaled by a factor $\beta_{\mathrm{scale}}$. We maintain a global phase $\phi(t)$ whose rate depends on $\delta_0(t)$:
\begin{equation}
  \phi(t + \Delta t) 
  \;=\; 
  \phi(t)
  \;+\;
  \bigl(1 + \alpha\,\delta_0(t)\bigr)\Delta t
  \label{eq:phase_accum_update}
\end{equation}
where $\Delta t$ is the simulation timestep, and $\alpha$ is a small gain modulating how strongly $\delta_0$ alters the nominal frequency. A larger $\delta_0(t)$ accelerates the gait cycle, whereas a negative value slows it down. Subsequently, a normalized phase:
\begin{equation}
  \varphi(t) 
  \;=\;
  \frac{\phi(t)}{T_{\mathrm{cycle}}},
\end{equation}
where $T_{\mathrm{cycle}}$ is the reference period. We then define two sinusoids:
\begin{equation}
  s(t) 
  \;=\; 
  \sin\,\!\Bigl(2\pi\,[\,\varphi(t) + \delta_1(t)\bigr]\Bigr),
\end{equation}
\begin{equation}
  c(t) 
  \;=\; 
  \sin\,\!\Bigl(2\pi\,[\,\varphi(t) + \delta_2(t) + \psi\bigr]\Bigr),
\end{equation}
where $\psi=0.5$ (by default) helps induce out-of-phase leg motions for a typical walking gait. Generally, $s(t)$ and $c(t)$ can be used to drive each foot's stance or swing trajectory, with $\delta_1(t)$ and $\delta_2(t)$ shifting their phases independently. As illustrated in Fig.~\ref{fig:gait}, the left circular diagram depicts how $\delta_0(t)$ affects the angular velocity of the global phase, while the right plot shows how $\delta_{1,2}(t)$ translate into shifted sine waves.

\subsubsection{Gait Phase Mask}
To classify stance versus swing for each foot, we apply the same threshold logic to whichever sinusoidal signal represents that foot’s phase, be it $s(t)$ or $c(t)$. For example, suppose we assign $s(t)$ to the left foot and $c(t)$ to the right foot. Then:
\begin{equation}
  \boldsymbol{m}_\mathrm{stance}(t)
  =
  \begin{bmatrix}
    \mathbb{I}\,\,\!\bigl\{\,s(t) \!\ge\! \gamma\bigr\},
    \quad
    \mathbb{I}\,\,\!\bigl\{\,c(t) \!\ge\! \gamma\bigr\}
  \end{bmatrix}^\top,
\end{equation}
where $\gamma$ is a small threshold (e.g.\ $0$ or $-0.1$) to form a dead zone that prevents rapid switching. Intuitively, if $s(t)$ (or $c(t)$) exceeds $\gamma$, the corresponding foot is designated \emph{stance}; if it lies below $\gamma$, we interpret the phase as \emph{swing}. Note that setting $\gamma = 0$ allows an immediate sign-based classification, while a non-zero $\gamma$ enforces a small overlap or double-support zone for smoother transitions. Though one could rely on a single function (e.g.\ $s(t)$) to control both feet, combining $s(t)$ and $c(t)$ enables more nuanced footfall timings or phase offsets when desired.

\vspace{2mm}
\noindent
\textbf{Remark.} By introducing three additional dimensions in the policy output---\(\delta_0, \delta_1,\) and \(\delta_2\)---and integrating them into the formulations described above [see Eq. \eqref{eq:phase_accum_update} and the defined sinusoids \(s(t),\,c(t)\)], this method can modulate both the global gait frequency and the phase offsets of each foot in real time. Consequently, the policy retains direct joint-control capabilities while flexibly adjusting stance-swing transitions, leading to robust and versatile bipedal gaits across diverse terrains and locomotion tasks.
         
\section{EXPERIMENTS}
\subsection{Experimental Platform}
\begin{figure}[h]
    \centering
    \includegraphics[width=0.5\textwidth]{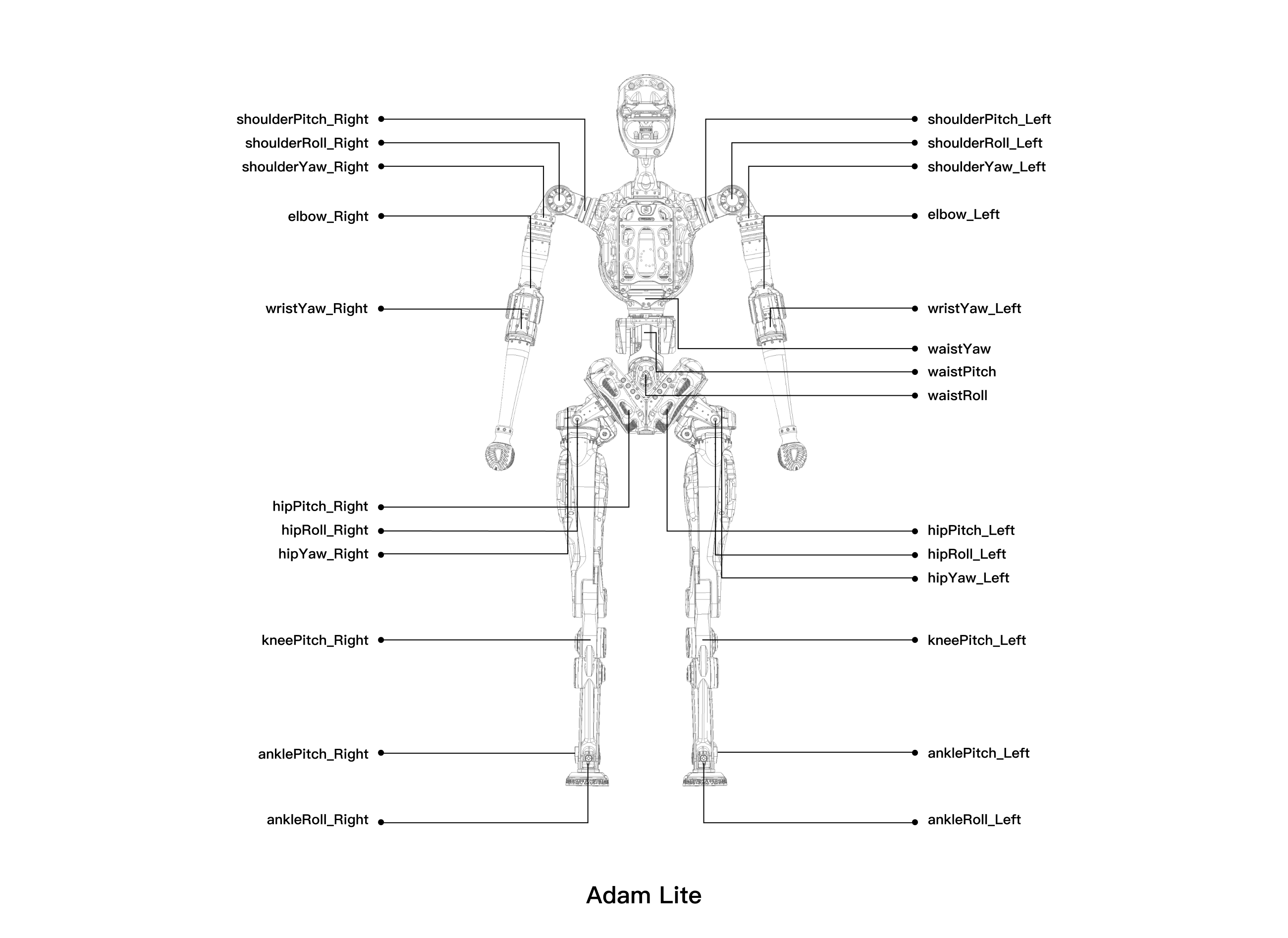}
    \caption{Schematic representation of the humanoid robot Adam Lite, showing the
front view.}
    \label{fig:robot_joint_lite}
\end{figure}

For our experiments, we utilize a full-sized humanoid robot known as ``Adam Lite''~\cite{adamlite}. As shown in Fig.~\ref{fig:robot_joint_lite}, Adam Lite stands at 1.6 meters tall, weighs approximately 60 kilograms, and features 25 degrees of freedom (DOF) distributed across its body. In practice, we only control 23 of these DOFs because the wrist yaw is not actively commanded. The robot employs QDD (quasi-direct drive) force-controlled actuators throughout its frame, with its legs equipped with four high-sensitivity, highly back-drivable actuators capable of generating up to 340\,N$\cdot$m of torque. The arms possess five degrees of freedom each, while the waist incorporates three. Its biomimetic joint configurations, particularly in the hip structure, enable human-like motion capabilities essential for our adaptive gait framework testing.

\subsubsection{Simulation Environment}
We conduct our training in the Isaac Gym~\cite{makoviychuk2021isaac} physics simulator, which enables massively parallel reinforcement learning for robotics. Our simulation environment consists of 4096 parallel instances, with the policy optimized by PPO~\cite{schulman2017proximal}. We create a detailed physics model of Adam Lite with accurate joint limits, mass distributions, and actuator characteristics to minimize the sim-to-real gap. Domain randomization is applied to physical parameters including mass properties ($\pm5\%$), center-of-mass position ($\pm5$\,cm), motor strength ($0.8$--$1.2\times$ default), and external disturbances through random impulses ($0$--$0.8$\,m/s) and forces ($\pm50$\,N). The high throughput of Isaac Gym accelerates training for our adaptive gait policies, allowing the process to complete in about 9 hours with 10k iterations on a single RTX~4090 device.

\begin{figure}[h]
    \centering
    \includegraphics[width=0.5\textwidth]{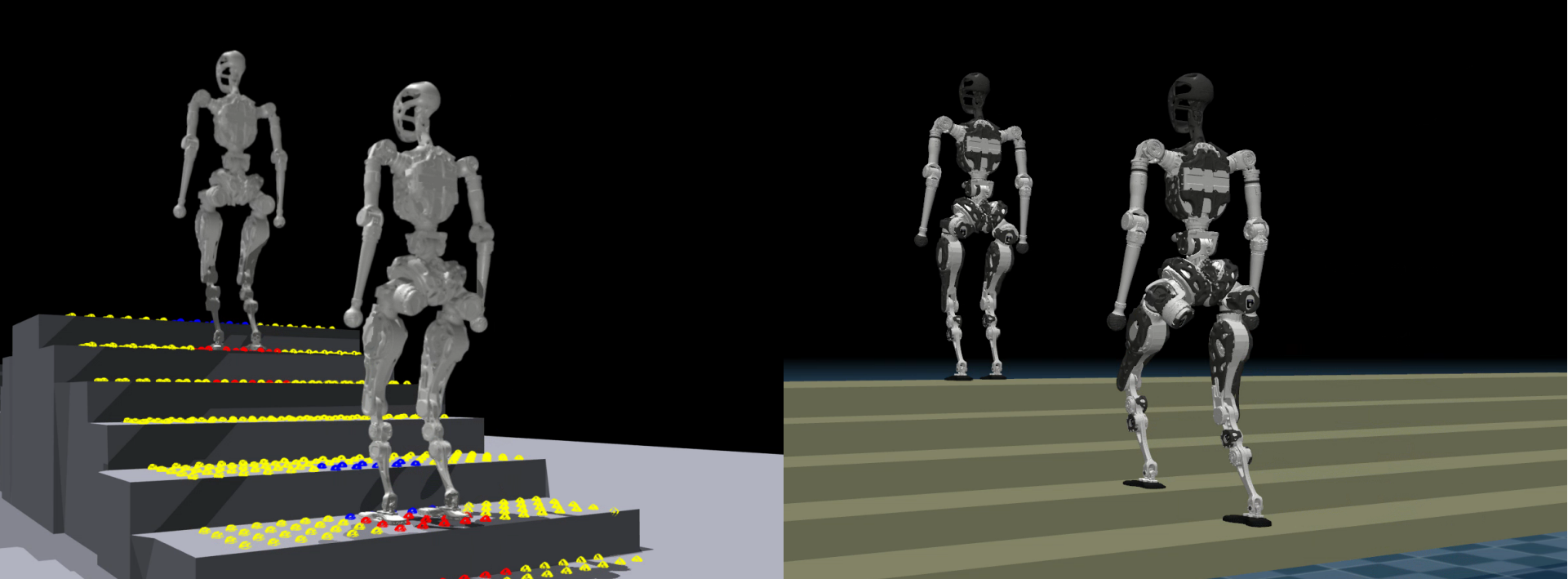}
    \caption{%
    Left: The humanoid robot in the Isaac Gym environment during training. 
    Right: The humanoid robot tested in the MuJoCo environment for validation.
    }
    \label{fig:isaac_mujoco}
\end{figure}

As depicted in Fig.~\ref{fig:isaac_mujoco}, the left panel shows the robot's training setup in Isaac Gym, while the right panel illustrates the same robotic model in MuJoCo. This dual-environment workflow facilitates reliable sim-to-sim transfer before progressing to real-world trials.

\subsubsection{Compared Methods}
To evaluate our method, we compare with several baselines and conduct ablation studies on key components:

\begin{itemize}
    \item \textbf{AdaptiveGait (Ours)}: Our approach.
    \item \textbf{AdaptiveGait w/ Sequential Contrast}: Sets the contrastive target to the next timestep's observation.
    \item \textbf{AdaptiveGait w/o Contrastive Learning}: Removes contrastive learning mechanism.
    \item \textbf{AdaptiveGait w/ Fixed Cycle}: Uses fixed cycle time and phase.
    \item \textbf{AdaptiveGait (Clock-free)}: Removes all clock-related inputs and action outputs.
    \item \textbf{DWL}~\cite{Gu-RSS-24}: Denoising World Model Learning.
    \item \textbf{DreamWaQ}~\cite{nahrendra2023dreamwaq}: Baseline with autoencoder for velocity estimation and a context vector.
\end{itemize}

\begin{figure}[h]
\centering
\includegraphics[width=0.5\textwidth]{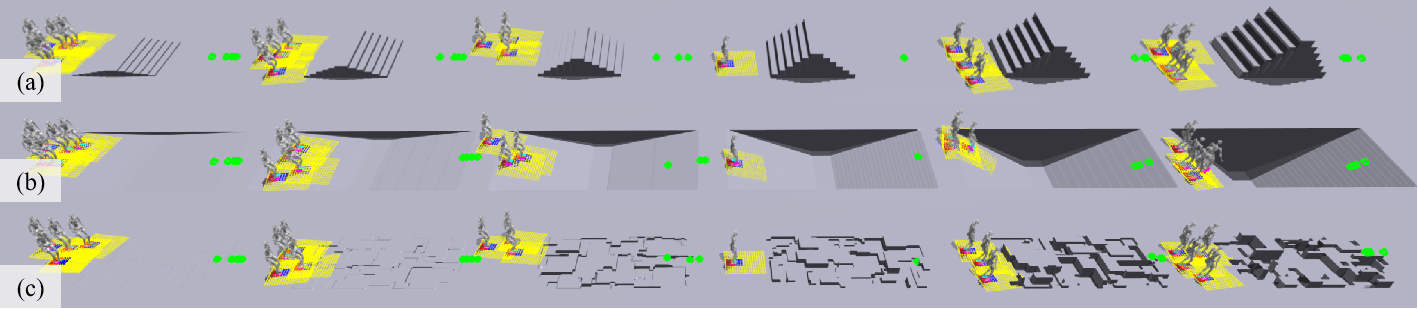}
\caption{Terrain curriculum: (a) Steps (5-35 cm), (b) Slopes (2-34°), (c) Uneven terrain. Robots must reach green targets without falling.}
\label{fig:sim_evaluation}
\end{figure}

For all experiments, we evaluate each method using three different random seeds and report the mean success rate with standard deviation. The evaluation uses a terrain curriculum as shown in Fig.~\ref{fig:sim_evaluation}, featuring (a) step obstacles (5-35 cm), (b) slopes (2-34°), and (c) uneven terrain. For each difficulty level, we deploy 200 robots and measure success rate as the percentage reaching green target points without falling.

\begin{figure}[h]
    \centering
    \includegraphics[width=0.5\textwidth]{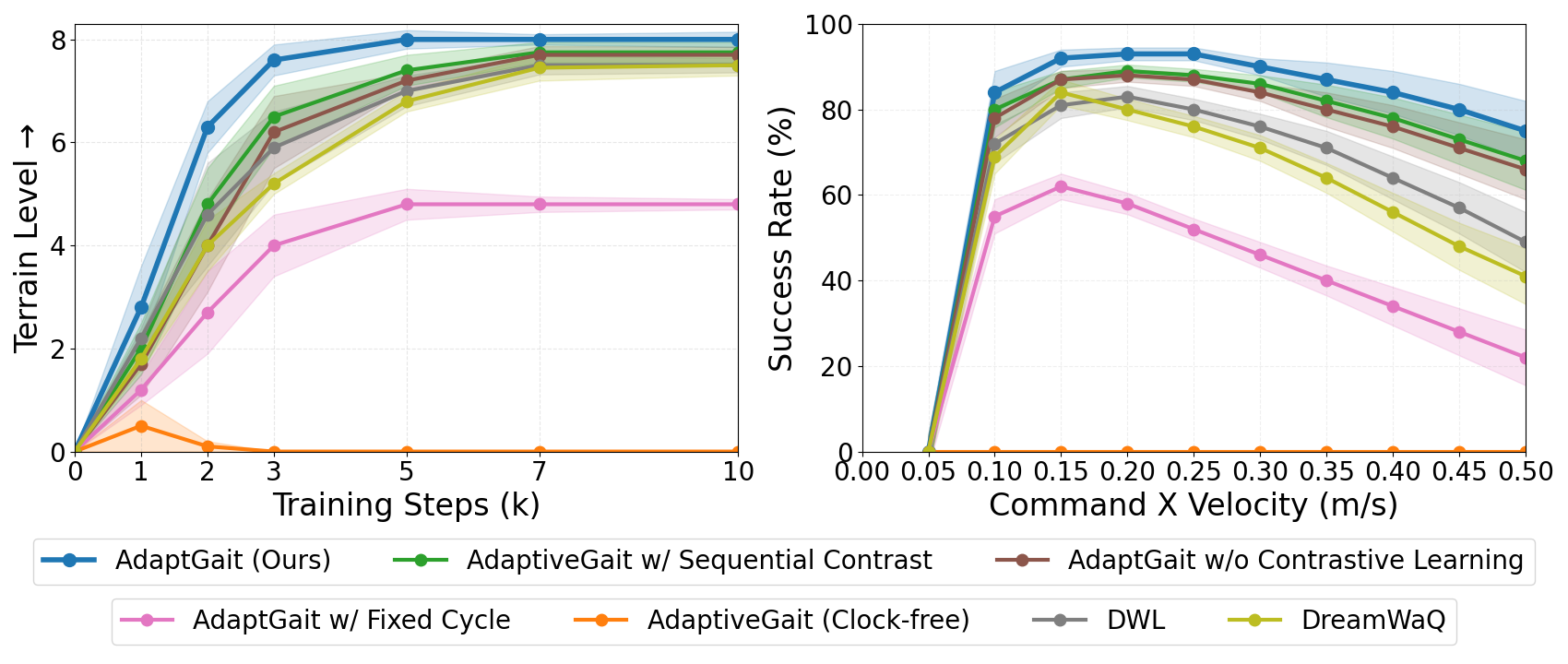}
    \caption{Left: Training progression showing terrain level advancement across different methods; Right: Success rate performance across varying command velocities.}
    \label{fig:training_performance}
\end{figure}

Fig.~\ref{fig:training_performance} demonstrates the effectiveness of our approach across different evaluation metrics. The left panel shows that AdaptiveGait (Ours) achieves the fastest terrain level progression during training, reaching the highest difficulty levels within just 5k training steps, significantly outpacing all baseline methods. This rapid progression indicates our method's superior learning efficiency on the challenging terrain course illustrated in Fig.~\ref{fig:sim_evaluation}. The right panel illustrates success rates across varying command velocities, where our method consistently maintains superior performance across the entire velocity range. Notably, all methods exhibit declining success rates as velocity increases, but AdaptiveGait (Ours) demonstrates the most robust performance degradation, maintaining above 75\% success rate even at the highest tested velocities. The ablation studies reveal that each component contributes meaningfully to the overall performance, with the terrain pathway and contrastive learning showing particularly significant impacts on success rates.

\subsubsection{Real-world Environment}
\begin{figure}[h]
    \centering
    \includegraphics[width=0.5\textwidth]{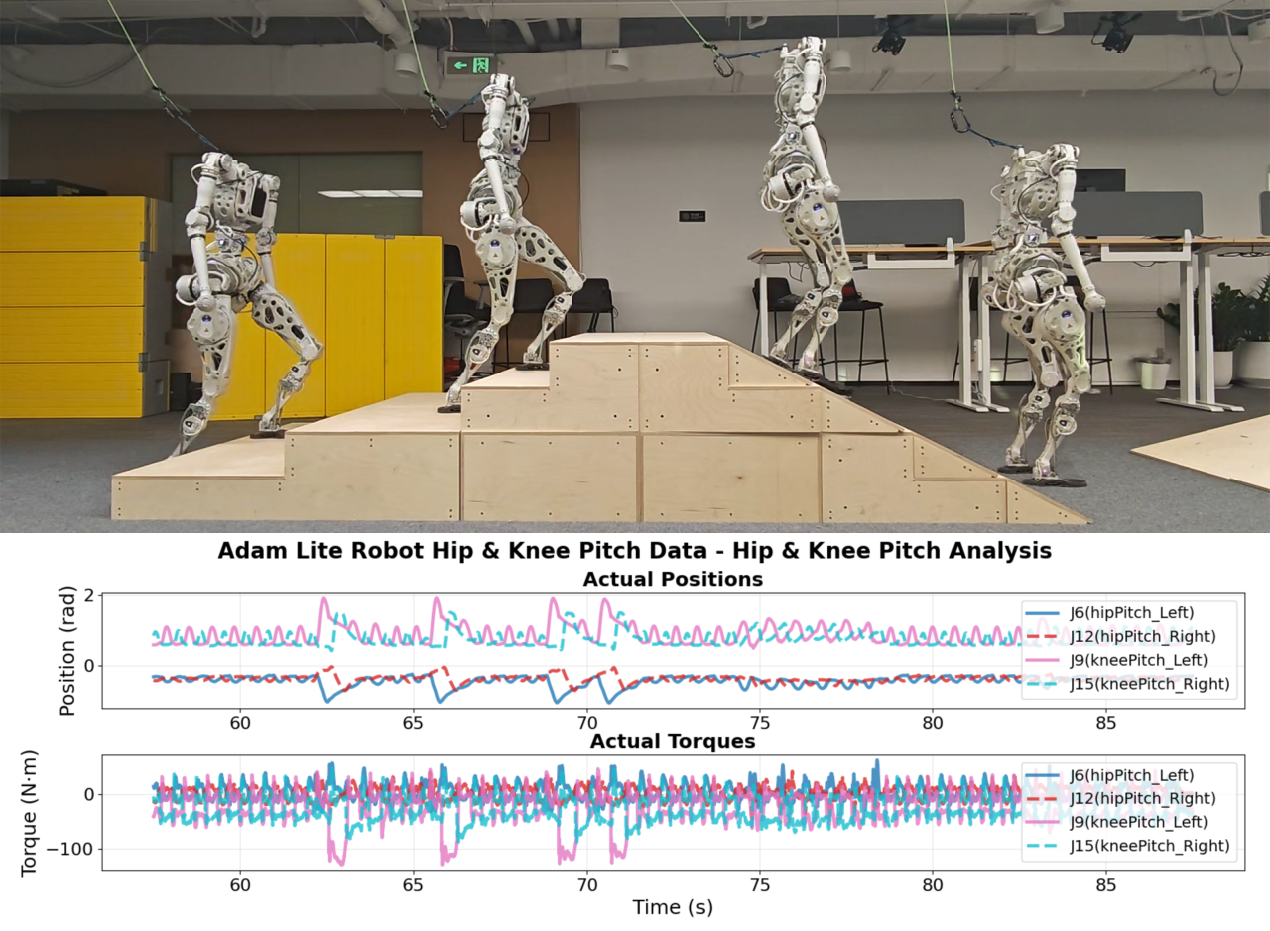}
    \caption{%
    \textbf{Top:} Adam Lite humanoid robot performing stair locomotion and slope traversal 
    in real-world testing environments.
    \textbf{Bottom:} Corresponding hip and knee joint position and torque data,
    demonstrating adaptive gait patterns and precise foot-lifting behavior.
    }
    \label{fig:upstairs}
\end{figure}

We evaluate our framework on challenging real-world terrains, including stairs, slopes, and uneven surfaces. Fig.~\ref{fig:upstairs} demonstrates the robot successfully ascending 15 cm stairs and descending 26.5° slopes, maintaining stable locomotion without backward slipping and exhibiting appropriate foot-lifting behavior when encountering vertical obstacles.

The joint trajectory analysis reveals distinct locomotion patterns across different terrain types. During stair ascent (62-72s), four pronounced peaks in hip and knee pitch positions correspond to precise leg-lifting motions for each step, while the relatively flat profiles between peaks indicate stable level-ground locomotion. A minor perturbation around 75s reflects the transition to slope descent, showcasing the robot's adaptive response to terrain changes. The torque data further illustrates the robot's energy-efficient control strategy. Higher torque variability during stair climbing reflects increased mechanical demands, while more consistent torque patterns on level terrain demonstrate optimized energy usage. This real-world validation confirms our method's ability to generate terrain-appropriate gaits that balance performance requirements with energy efficiency.


\begin{figure}[h]
    \centering
    \includegraphics[width=0.5\textwidth]{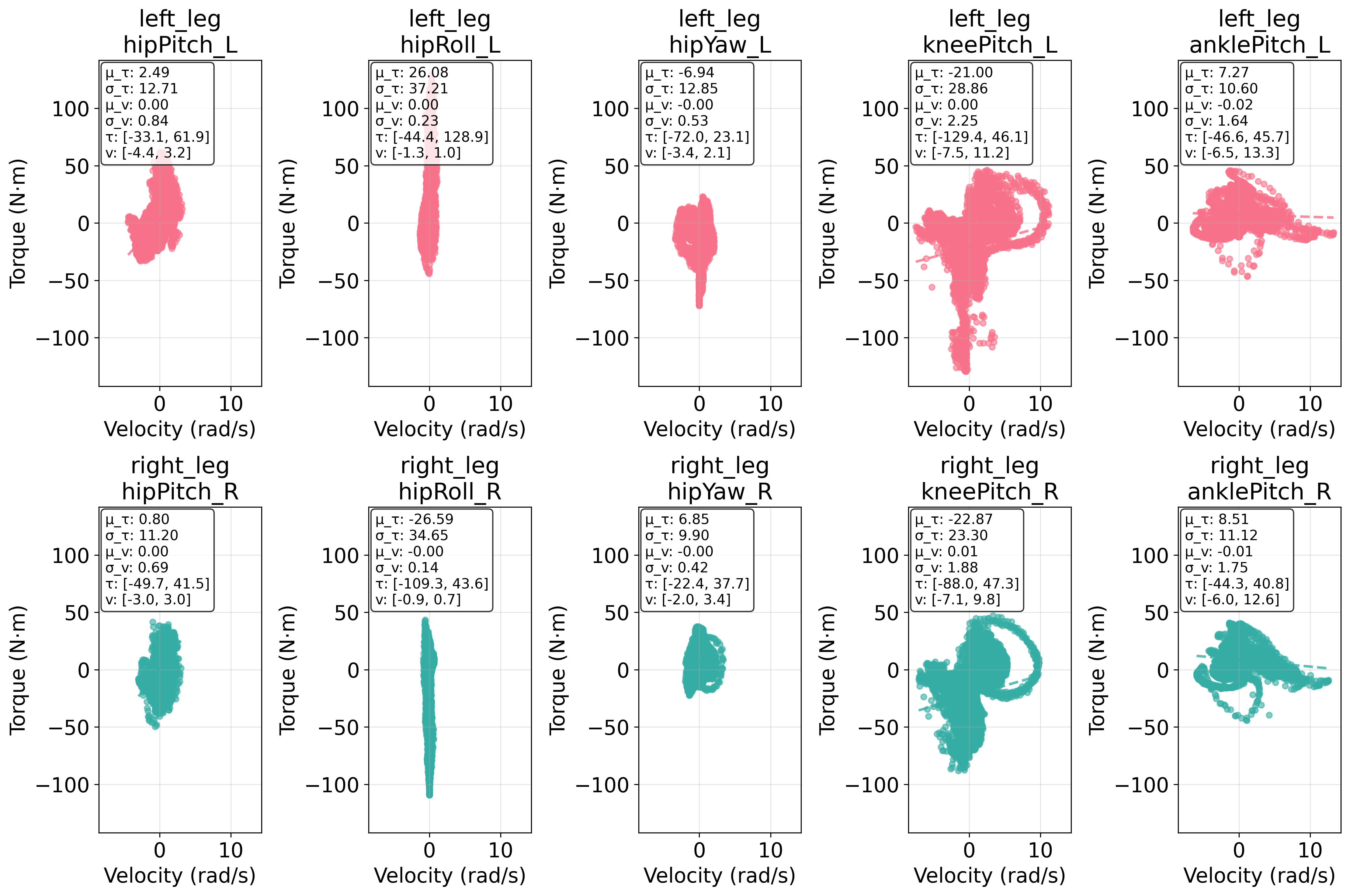}
    \caption{%
    Adam Lite humanoid robot climbing stairs, 
    illustrating scatter plots of joint velocity vs.\ torque 
    under stair-climbing conditions. 
    }
    \label{fig:scatter_stair}
\end{figure}

Fig.~\ref{fig:scatter_stair} shows the velocity-torque scatter plots for the major leg joints as the robot ascends stairs. Notably, the hip pitch (both left and right) and knee pitch (both left and right) demand the highest torque, underlining the significant effort required to lift the robot's body mass onto each step. These high-torque peaks correlate with lower joint velocities when the joint provides forceful support, reflecting the policy's learned trade-off between torque output and velocity regulation during stair climbing. This data highlights the efficacy of our approach in managing the large joint torques necessary for steep inclines and vertical transitions, without sacrificing stability or smoothness.

\begin{figure}[h]
    \centering
    \includegraphics[width=0.5\textwidth]{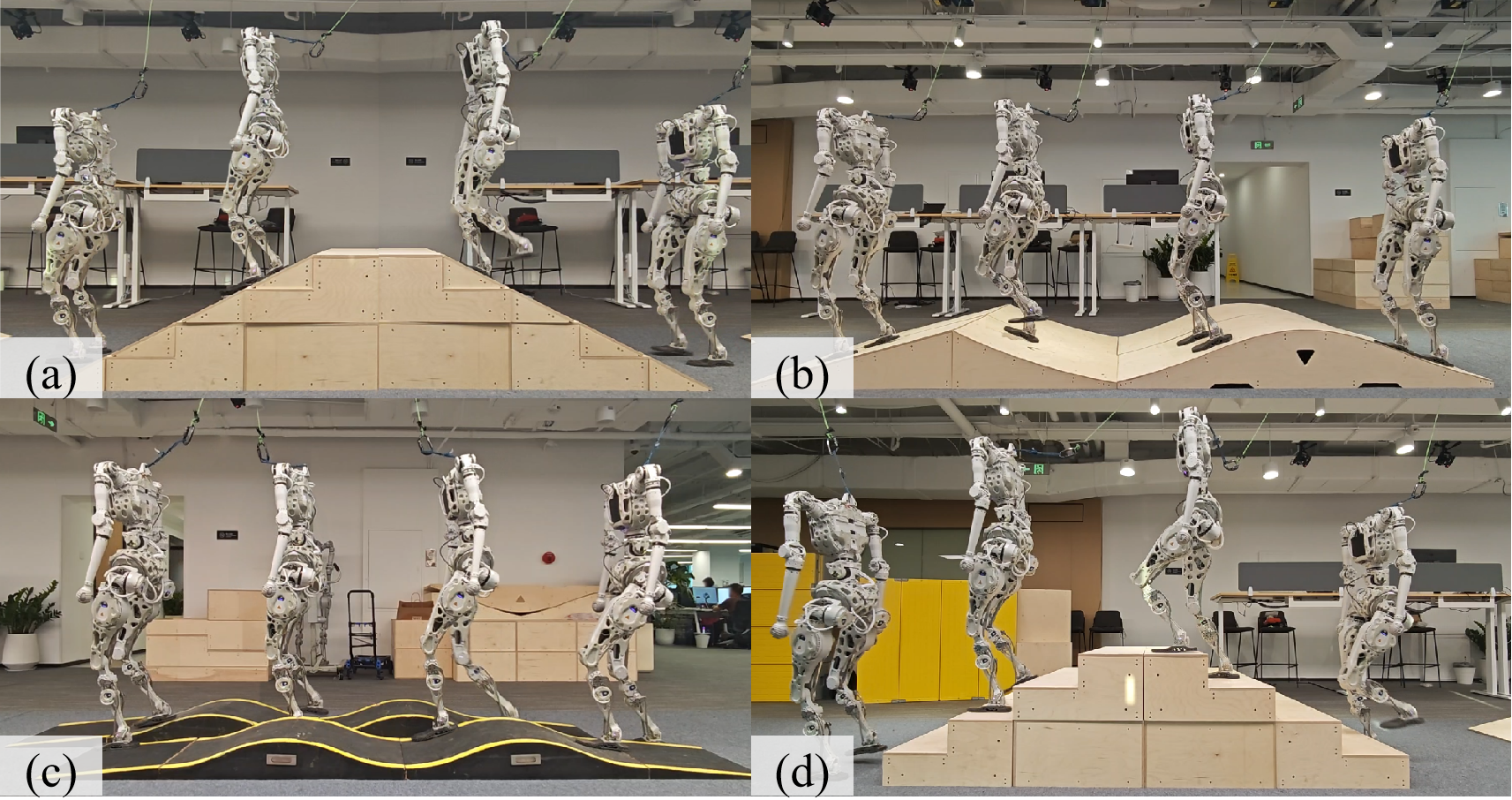}
    \caption{%
    Adam Lite humanoid robot locomotion on various terrains:
    (a) 26.5° slope \emph{(top-left)},
    (b) 14.7°, 20.5°, and 16.7° slopes from left to right \emph{(top-right)},
    (c) uneven terrain \emph{(bottom-left)},
    and (d) 15\,cm steps \emph{(bottom-right)}.%
    }
    \label{fig:stairs_and_slopes}
\end{figure}

\begin{figure}[h]
    \centering
    \includegraphics[width=0.5\textwidth]{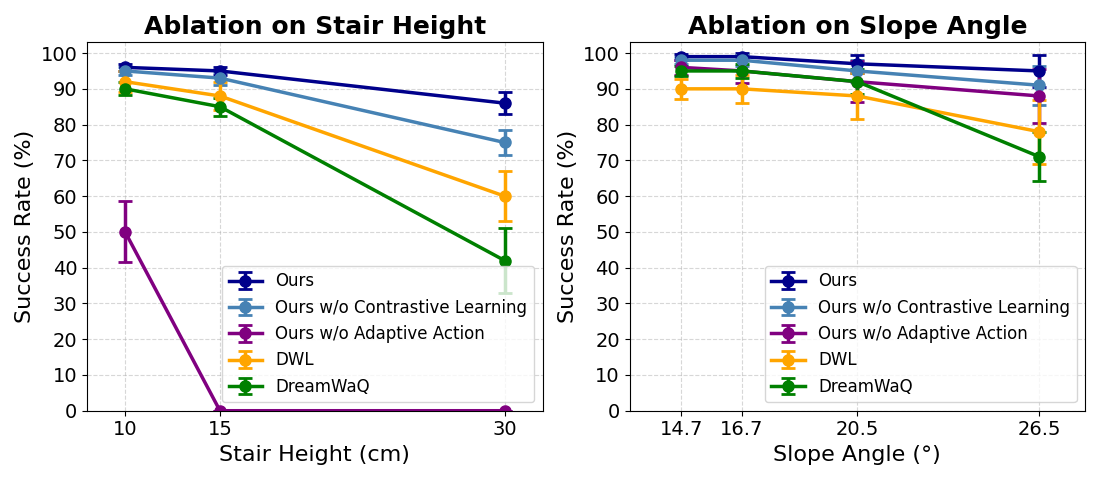}
    \caption{Ablation studies on terrain-specific challenges. Left: Success rates across varying stair heights; Right: Performance on different slope angles. All experiments conducted at 0.15 m/s forward velocity.}
    \label{fig:ablation}
\end{figure}

\subsubsection{Terrain-Specific Ablation Analysis}

Our method demonstrates exceptional robustness across diverse challenging terrains in real-world deployment. Fig.~\ref{fig:frame} showcases how the Adam Lite humanoid robot successfully locomotes over step obstacles up to 30 cm in height, while Fig.~\ref{fig:stairs_and_slopes} illustrates performance across four distinct terrain types: steep slopes up to 26.5°, multi-slope sequences with varying angles (14.7°, 20.5°, and 16.7°), uneven terrain, and 15 cm step traversal. These real-world demonstrations validate our approach's practical effectiveness across the full spectrum of challenging locomotion scenarios.

To quantitatively evaluate this performance, Fig.~\ref{fig:ablation} presents comprehensive ablation studies where all methods are evaluated at 0.15 m/s forward velocity. Each experiment was conducted with 25 trials per policy using three different training seeds. For stair traversing (left panel), our method maintains over 95\% success rate on moderate heights (10-15 cm) and achieves 85\% success even on challenging 30 cm stairs. Removing contrastive learning significantly impacts performance on higher stairs, while baseline methods DWL and DreamWAQ show substantial drops, falling below 60\% on extreme scenarios.

On slope traversal (right panel), our approach demonstrates consistent superiority across all tested angles, maintaining over 90\% success rate even on steep 26.5° slopes. The ablation studies confirm that each component contributes meaningfully to terrain adaptability, with our complete framework showing the most robust performance degradation patterns.

\section{CONCLUSIONS}

We present a unified approach to adaptive humanoid locomotion that successfully combines architectural innovations with adaptive control mechanisms. Our asymmetric actor-critic framework with contrastive learning effectively bridges the gap between simulation training and real-world deployment, while the adaptive gait cycle mechanism enables dynamic terrain adaptation without sacrificing stability.

Experimental validation on the Adam Lite humanoid robot demonstrated the practical effectiveness of our approach, successfully achieving locomotion over challenging terrains including stairs up to 30 cm high and slopes up to 26.5° with superior robustness compared to existing baselines. Our framework advances superior results in humanoid locomotion by proving that principled architectural design can achieve both training efficiency and deployment robustness for real-world applications.










\bibliographystyle{IEEEtran}
\bibliography{reference}

\begin{thebibliography}{10}
\providecommand{\url}[1]{#1}
\csname url@samestyle\endcsname
\providecommand{\newblock}{\relax}
\providecommand{\bibinfo}[2]{#2}
\providecommand{\BIBentrySTDinterwordspacing}{\spaceskip=0pt\relax}
\providecommand{\BIBentryALTinterwordstretchfactor}{4}
\providecommand{\BIBentryALTinterwordspacing}{\spaceskip=\fontdimen2\font plus
\BIBentryALTinterwordstretchfactor\fontdimen3\font minus \fontdimen4\font\relax}
\providecommand{\BIBforeignlanguage}[2]{{%
\expandafter\ifx\csname l@#1\endcsname\relax
\typeout{** WARNING: IEEEtran.bst: No hyphenation pattern has been}%
\typeout{** loaded for the language `#1'. Using the pattern for}%
\typeout{** the default language instead.}%
\else
\language=\csname l@#1\endcsname
\fi
#2}}
\providecommand{\BIBdecl}{\relax}
\BIBdecl

\bibitem{siekmann2021sim}
J.~Siekmann, Y.~Godse, A.~Fern, and J.~Hurst, ``Sim-to-real learning of all common bipedal gaits via periodic reward composition,'' in \emph{2021 IEEE International Conference on Robotics and Automation (ICRA)}.\hskip 1em plus 0.5em minus 0.4em\relax IEEE, 2021, pp. 7309--7315.

\bibitem{siekmann2021blind}
J.~Siekmann, K.~Green, J.~Warila, A.~Fern, and J.~Hurst, ``Blind bipedal stair traversal via sim-to-real reinforcement learning,'' \emph{arXiv preprint arXiv:2105.08328}, 2021.

\bibitem{xie2020learning}
Z.~Xie, P.~Clary, J.~Dao, P.~Morais, J.~Hurst, and M.~Panne, ``Learning locomotion skills for cassie: Iterative design and sim-to-real,'' in \emph{Conference on Robot Learning}.\hskip 1em plus 0.5em minus 0.4em\relax PMLR, 2020, pp. 317--329.

\bibitem{xie2018feedback}
Z.~Xie, G.~Berseth, P.~Clary, J.~Hurst, and M.~Van~de Panne, ``Feedback control for cassie with deep reinforcement learning,'' in \emph{2018 IEEE/RSJ International Conference on Intelligent Robots and Systems (IROS)}.\hskip 1em plus 0.5em minus 0.4em\relax IEEE, 2018, pp. 1241--1246.

\bibitem{rodriguez2021deepwalk}
D.~Rodriguez and S.~Behnke, ``Deepwalk: Omnidirectional bipedal gait by deep reinforcement learning,'' in \emph{2021 IEEE international conference on robotics and automation (ICRA)}.\hskip 1em plus 0.5em minus 0.4em\relax IEEE, 2021, pp. 3033--3039.

\bibitem{li2021reinforcement}
Z.~Li, X.~Cheng, X.~B. Peng, P.~Abbeel, S.~Levine, G.~Berseth, and K.~Sreenath, ``Reinforcement learning for robust parameterized locomotion control of bipedal robots,'' in \emph{2021 IEEE International Conference on Robotics and Automation (ICRA)}.\hskip 1em plus 0.5em minus 0.4em\relax IEEE, 2021, pp. 2811--2817.

\bibitem{li2023robust}
Z.~Li, X.~B. Peng, P.~Abbeel, S.~Levine, G.~Berseth, and K.~Sreenath, ``Robust and versatile bipedal jumping control through multi-task reinforcement learning,'' \emph{arXiv preprint arXiv:2302.09450}, vol.~1, 2023.

\bibitem{li2024reinforcement}
Z.~Li, X.~B. Peng, P.~Abbeel, S.~Levine, G.~Berseth \emph{et~al.}, ``Reinforcement learning for versatile, dynamic, and robust bipedal locomotion control,'' \emph{The International Journal of Robotics Research}, p. 02783649241285161, 2024.

\bibitem{duan2022learning}
H.~Duan, A.~Malik, M.~S. Gadde, J.~Dao, A.~Fern, and J.~Hurst, ``Learning dynamic bipedal walking across stepping stones,'' in \emph{2022 IEEE/RSJ International Conference on Intelligent Robots and Systems (IROS)}.\hskip 1em plus 0.5em minus 0.4em\relax IEEE, 2022, pp. 6746--6752.

\bibitem{radosavovic2024real}
I.~Radosavovic, T.~Xiao, B.~Zhang, T.~Darrell, J.~Malik, and K.~Sreenath, ``Real-world humanoid locomotion with reinforcement learning,'' \emph{Science Robotics}, vol.~9, no.~89, p. eadi9579, 2024.

\bibitem{duan2024learning}
H.~Duan, B.~Pandit, M.~S. Gadde, B.~Van~Marum, J.~Dao, C.~Kim, and A.~Fern, ``Learning vision-based bipedal locomotion for challenging terrain,'' in \emph{2024 IEEE International Conference on Robotics and Automation (ICRA)}, 2024, pp. 56--62.

\bibitem{zhuang2024humanoid}
\BIBentryALTinterwordspacing
Z.~Zhuang, S.~Yao, and H.~Zhao, ``Humanoid parkour learning,'' in \emph{8th Annual Conference on Robot Learning}, 2024. [Online]. Available: \url{https://openreview.net/forum?id=fs7ia3FqUM}
\BIBentrySTDinterwordspacing

\bibitem{allshire2025visual}
A.~Allshire, H.~Choi, J.~Zhang, D.~McAllister, A.~Zhang, C.~M. Kim, T.~Darrell, P.~Abbeel, J.~Malik, and A.~Kanazawa, ``Visual imitation enables contextual humanoid control,'' \emph{arXiv preprint arXiv:2505.03729}, 2025.

\bibitem{wang2025moremixtureresidualexperts}
\BIBentryALTinterwordspacing
D.~Wang, X.~Wang, X.~Liu, J.~Shi, Y.~Zhao, C.~Bai, and X.~Li, ``More: Mixture of residual experts for humanoid lifelike gaits learning on complex terrains,'' 2025. [Online]. Available: \url{https://arxiv.org/abs/2506.08840}
\BIBentrySTDinterwordspacing

\bibitem{adamlite}
``\textcolor{black}{Adam Lite},'' \url{https://www.pndbotics.com/humanoid}, \textcolor{black}{2025}, \textcolor{black}{Accessed: March 26, 2025}.

\bibitem{ijspeert2008central}
A.~J. Ijspeert, ``Central pattern generators for locomotion control in animals and robots: a review,'' \emph{Neural networks}, vol.~21, no.~4, pp. 642--653, 2008.

\bibitem{li2024ai}
G.~Li, A.~Ijspeert, and M.~Hayashibe, ``Ai-cpg: Adaptive imitated central pattern generators for bipedal locomotion learned through reinforced reflex neural networks,'' \emph{IEEE Robotics and Automation Letters}, 2024.

\bibitem{zhang2021adaptive}
B.~Zhang, M.~Zhou, W.~Xu \emph{et~al.}, ``An adaptive framework of real-time continuous gait phase variable estimation for lower-limb wearable robots,'' \emph{Robotics and Autonomous Systems}, vol. 143, p. 103842, 2021.

\bibitem{kajita2003biped}
S.~Kajita, F.~Kanehiro, K.~Kaneko, K.~Fujiwara, K.~Harada, K.~Yokoi, and H.~Hirukawa, ``Biped walking pattern generation by using preview control of zero-moment point,'' in \emph{2003 IEEE international conference on robotics and automation (Cat. No. 03CH37422)}, vol.~2.\hskip 1em plus 0.5em minus 0.4em\relax IEEE, 2003, pp. 1620--1626.

\bibitem{kajita20013d}
S.~Kajita, F.~Kanehiro, K.~Kaneko, K.~Yokoi, and H.~Hirukawa, ``The 3d linear inverted pendulum mode: A simple modeling for a biped walking pattern generation,'' in \emph{Proceedings 2001 IEEE/RSJ International Conference on Intelligent Robots and Systems. Expanding the Societal Role of Robotics in the the Next Millennium (Cat. No. 01CH37180)}, vol.~1.\hskip 1em plus 0.5em minus 0.4em\relax IEEE, 2001, pp. 239--246.

\bibitem{wieber2006trajectory}
P.-B. Wieber, ``Trajectory free linear model predictive control for stable walking in the presence of strong perturbations,'' in \emph{2006 6th IEEE-RAS International Conference on Humanoid Robots}.\hskip 1em plus 0.5em minus 0.4em\relax IEEE, 2006, pp. 137--142.

\bibitem{englsberger2013three}
J.~Englsberger, C.~Ott, and A.~Albu-Sch{\"a}ffer, ``Three-dimensional bipedal walking control using divergent component of motion,'' in \emph{2013 IEEE/RSJ International Conference on Intelligent Robots and Systems}.\hskip 1em plus 0.5em minus 0.4em\relax IEEE, 2013, pp. 2600--2607.

\bibitem{xie2021compliant}
S.~Xie, X.~Li, H.~Zhong, C.~Hu, and L.~Gao, ``Compliant bipedal walking based on variable spring-loaded inverted pendulum model with finite-sized foot,'' in \emph{2021 6th IEEE International Conference on Advanced Robotics and Mechatronics (ICARM)}.\hskip 1em plus 0.5em minus 0.4em\relax IEEE, 2021, pp. 667--672.

\bibitem{peng2017deeploco}
X.~B. Peng, G.~Berseth, K.~Yin, and M.~Van De~Panne, ``Deeploco: Dynamic locomotion skills using hierarchical deep reinforcement learning,'' \emph{Acm transactions on graphics (tog)}, vol.~36, no.~4, pp. 1--13, 2017.

\bibitem{gu2024humanoid}
X.~Gu, Y.-J. Wang, and J.~Chen, ``Humanoid-gym: Reinforcement learning for humanoid robot with zero-shot sim2real transfer,'' \emph{arXiv preprint arXiv:2404.05695}, 2024.

\bibitem{xue2025unified}
Y.~Xue, W.~Dong, M.~Liu, W.~Zhang, and J.~Pang, ``A unified and general humanoid whole-body controller for fine-grained locomotion,'' in \emph{Robotics: Science and Systems (RSS)}, 2025.

\bibitem{long2025learning}
J.~Long, J.~Ren, M.~Shi, Z.~Wang, T.~Huang, P.~Luo, and J.~Pang, ``Learning humanoid locomotion with perceptive internal model,'' in \emph{2025 IEEE International Conference on Robotics and Automation (ICRA)}.\hskip 1em plus 0.5em minus 0.4em\relax IEEE, 2025, pp. 9997--10\,003.

\bibitem{he2025attention}
J.~He, C.~Zhang, F.~Jenelten, R.~Grandia, M.~B{\"a}cher, and M.~Hutter, ``Attention-based map encoding for learning generalized legged locomotion,'' \emph{Science Robotics}, vol.~10, no. 105, p. eadv3604, 2025.

\bibitem{wang2025beamdojo}
H.~Wang, Z.~Wang, J.~Ren, Q.~Ben, T.~Huang, W.~Zhang, and J.~Pang, ``Beamdojo: Learning agile humanoid locomotion on sparse footholds,'' in \emph{Robotics: Science and Systems ({RSS})}, 2025.

\bibitem{kumar2022adapting}
A.~Kumar, Z.~Li, J.~Zeng, D.~Pathak, K.~Sreenath, and J.~Malik, ``Adapting rapid motor adaptation for bipedal robots,'' in \emph{2022 IEEE/RSJ International Conference on Intelligent Robots and Systems (IROS)}.\hskip 1em plus 0.5em minus 0.4em\relax IEEE, 2022, pp. 1161--1168.

\bibitem{Gu-RSS-24}
X.~Gu, Y.-J. Wang, X.~Zhu, C.~Shi, Y.~Guo, Y.~Liu, and J.~Chen, ``{Advancing Humanoid Locomotion: Mastering Challenging Terrains with Denoising World Model Learning},'' in \emph{Proceedings of Robotics: Science and Systems}, Delft, Netherlands, July 2024.

\bibitem{sun2025learning}
W.~Sun, L.~Chen, Y.~Su, B.~Cao, Y.~Liu, and Z.~Xie, ``Learning humanoid locomotion with world model reconstruction,'' \emph{arXiv preprint arXiv:2502.16230}, 2025.

\bibitem{long2023hybrid}
J.~Long, Z.~Wang, Q.~Li, J.~Gao, L.~Cao, and J.~Pang, ``Hybrid internal model: Learning agile legged locomotion with simulated robot response,'' \emph{arXiv preprint arXiv:2312.11460}, 2023.

\bibitem{spaan2012partially}
M.~T. Spaan, ``Partially observable markov decision processes,'' in \emph{Reinforcement learning: State-of-the-art}.\hskip 1em plus 0.5em minus 0.4em\relax Springer, 2012, pp. 387--414.

\bibitem{schulman2017proximal}
J.~Schulman, F.~Wolski, P.~Dhariwal, A.~Radford, and O.~Klimov, ``Proximal policy optimization algorithms,'' \emph{arXiv preprint arXiv:1707.06347}, 2017.

\bibitem{makoviychuk2021isaac}
V.~Makoviychuk, L.~Wawrzyniak, Y.~Guo, M.~Lu, K.~Storey, M.~Macklin, D.~Hoeller, N.~Rudin, A.~Allshire, A.~Handa \emph{et~al.}, ``Isaac gym: High performance gpu-based physics simulation for robot learning,'' \emph{arXiv preprint arXiv:2108.10470}, 2021.

\bibitem{nahrendra2023dreamwaq}
I.~M.~A. Nahrendra, B.~Yu, and H.~Myung, ``Dreamwaq: Learning robust quadrupedal locomotion with implicit terrain imagination via deep reinforcement learning,'' in \emph{2023 IEEE International Conference on Robotics and Automation (ICRA)}.\hskip 1em plus 0.5em minus 0.4em\relax IEEE, 2023, pp. 5078--5084.

\end{thebibliography}
 
\end{document}